\documentclass[twocolumn,10pt]{article}

\usepackage{fullpage}
\usepackage{amsmath,amssymb,amsfonts}
\usepackage{graphicx,epsfig,psfrag}
\input{epsf}

\newcommand{\R}{\mathbb{R}}       
\newcommand{\ssp}{\hspace{0.5pt}}           
\newcommand{\bssp}{\hspace{-0.5pt}}         
\renewcommand{\P}{\mathbb{P}}         
\newcommand{\eref}[1]{(\ref{#1})}       

\begin{document}

\title{Geometric Models of Rolling-Shutter Cameras}
\author{Marci Meingast, Christopher Geyer, and Shankar Sastry}
\texttt{\{marci,cgeyer,sastry\}@eecs.berkeley.edu\\
EECS Department, University of California, Berkeley}
\date{} \maketitle

\begin{abstract}
Cameras with rolling shutters are becoming more common as
low-power, low-cost CMOS sensors are being used more frequently in
cameras. The rolling shutter means that not all scanlines are
exposed over the same time interval.  The effects of a rolling
shutter are noticeable when either the camera or objects in the
scene are moving and can lead to systematic biases in projection
estimation.  We develop a general projection equation for a
rolling shutter camera and show how it is affected by different
types of camera motion.  In the case of fronto-parallel motion, we
show how that camera can be modeled as an X-slit camera.  We also
develop approximate projection equations for a non-zero angular
velocity about the optical axis and approximate the projection
equation for a constant velocity screw motion. We demonstrate how
the rolling shutter effects the projective geometry of the camera
and in turn the structure-from-motion.
\end{abstract}

\section{Introduction}

The motivation for this paper is the use of inexpensive camera sensors
for robotics.  Maintaining accuracy in this low-cost realm is
difficult to do, especially when sacrifices in sensor design are made
in order to reduce cost or power consumption.  For example, as of
2004, many of the CMOS chips manufactured by OEMs and integrated into
off-the-shelf cameras do not have a global
shutter.\footnote{Fortunately, some progress has been made in CMOS
chips with a global shutter, e.g. see \cite{wany03ted}.}  Unlike CCD
chips with interline transfer, the individual pixels in a CMOS camera
chip typically cannot hold and store intensities, forcing a so-called
\emph{rolling shutter} whereby each scanline is exposed, read out, and
in the case of most Firewire cameras, is immediately transmitted to
the host computer.  Furthermore, for cameras connected via a Firewire
bus, the digital camera specification \cite{IIDC} requires that cameras
distribute data over a constant number of packets regardless of the
framerate.  Thus, if the camera has no on-board buffer, the overall
exposure for a single frame is inversely proportional to the
framerate.

\begin{figure}[htb]
\centerline{
\begin{tabular}{cc}
\includegraphics[scale=0.257]{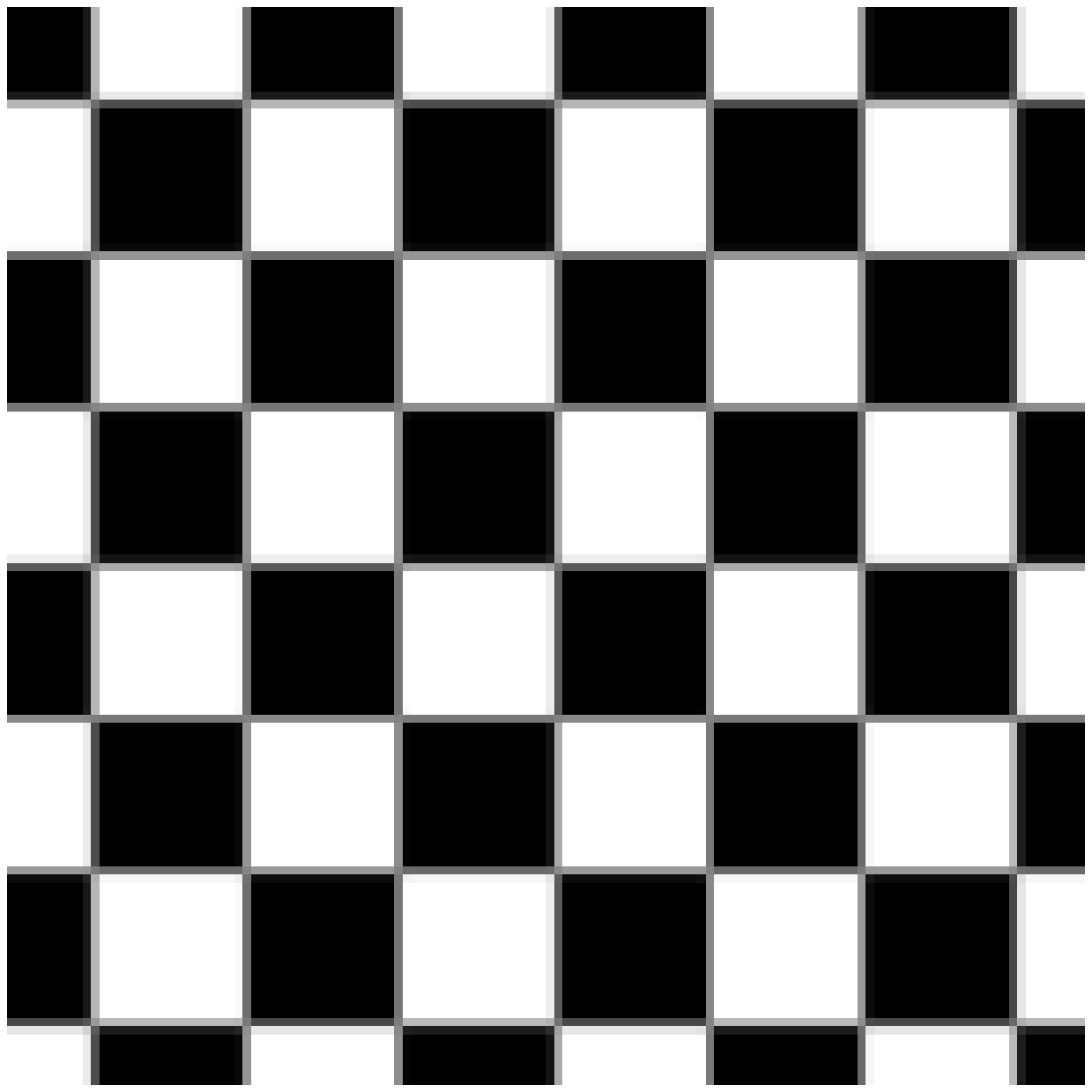} &
\includegraphics[scale=0.257]{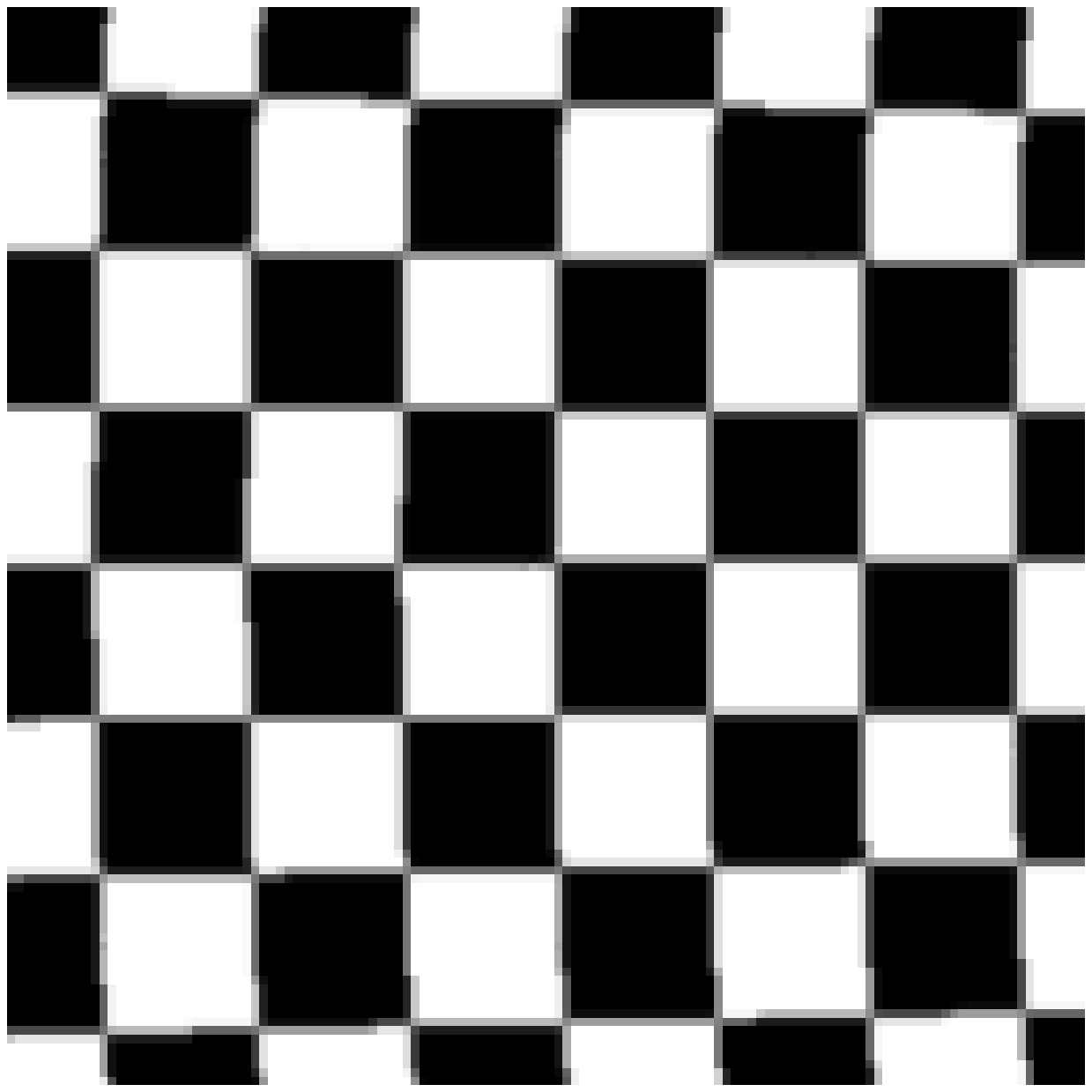} \\
\includegraphics[scale=0.257]{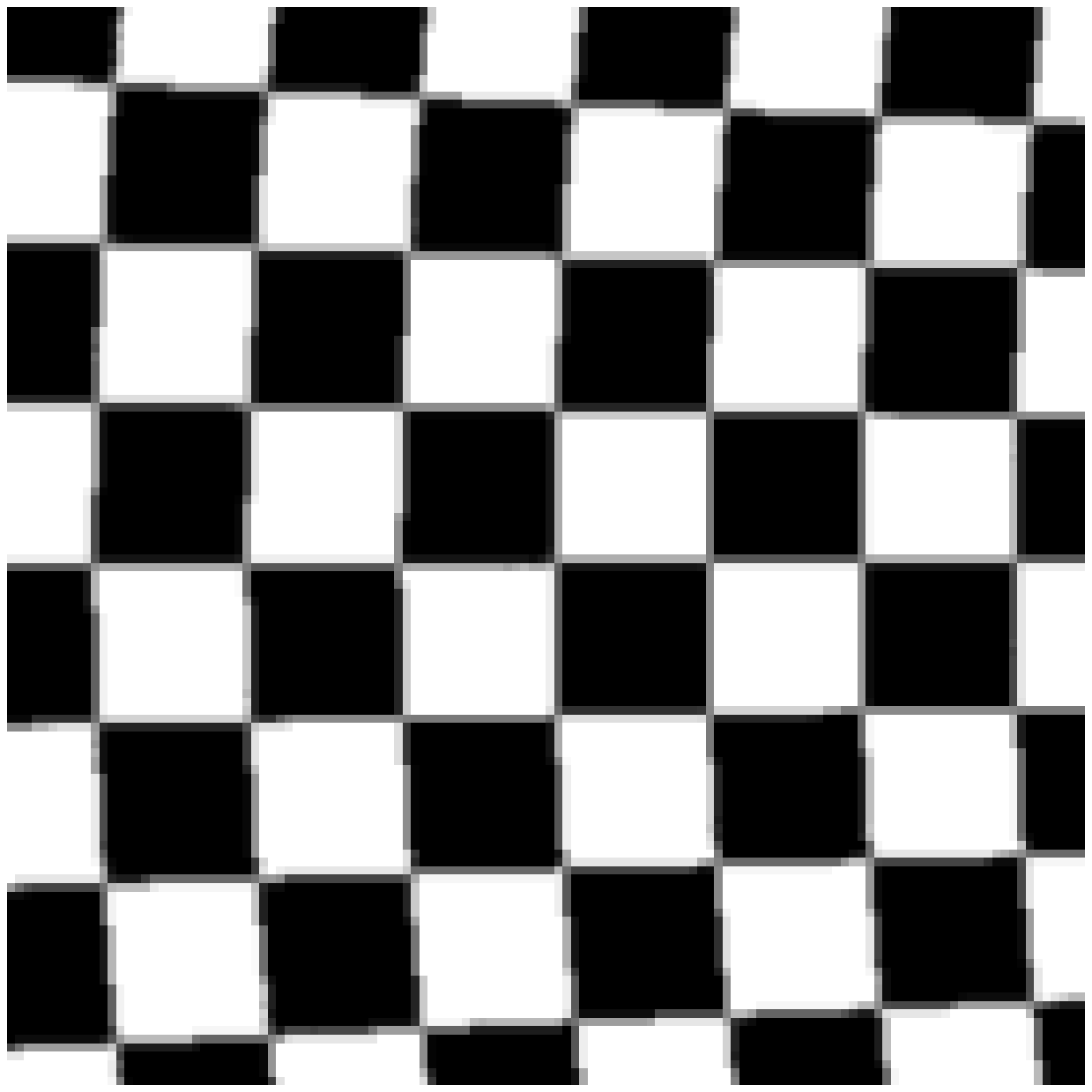} &
\includegraphics[scale=0.257]{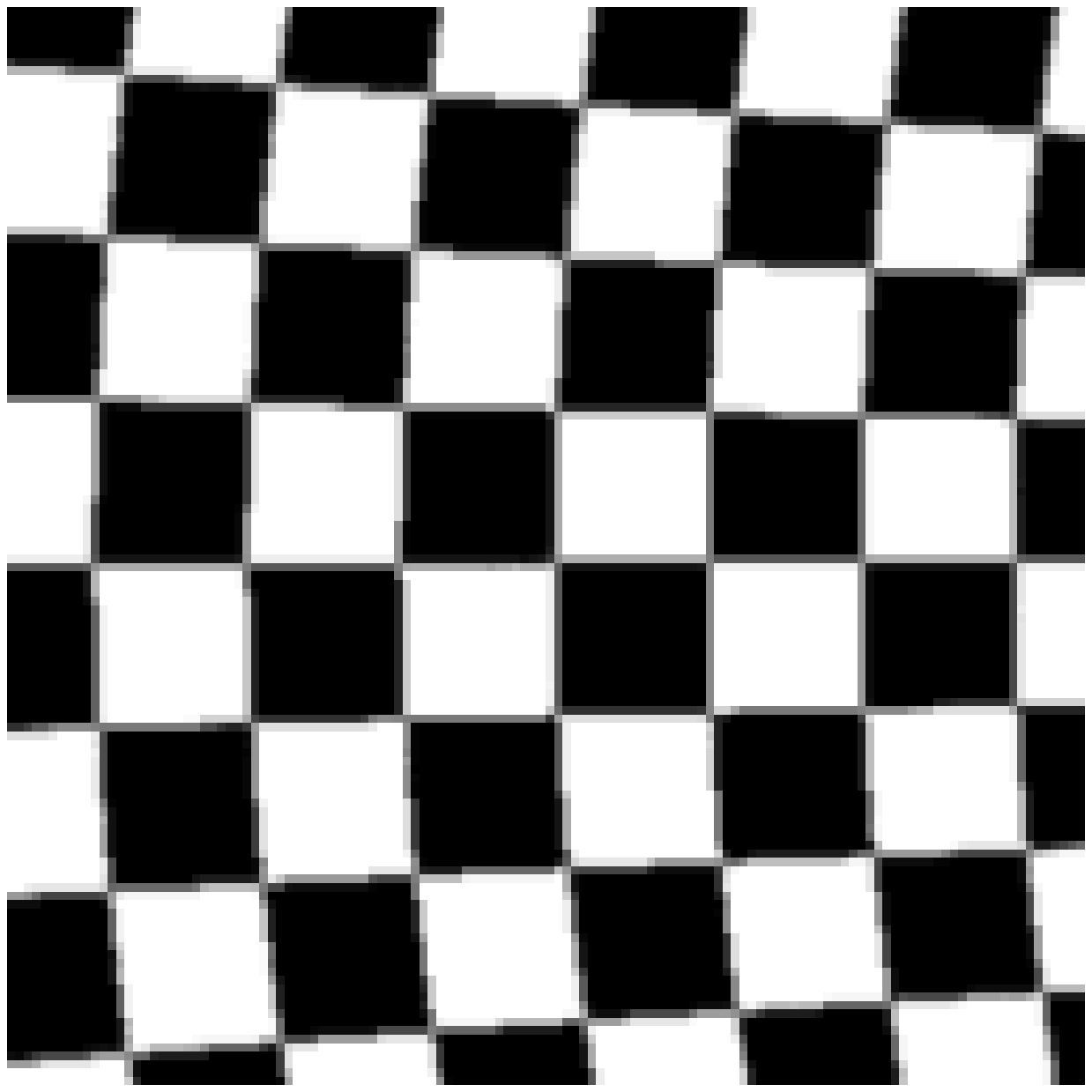}
\end{tabular}}
\caption{\label{checkboard} These four images show a physically
rectilinear checkerboard as seen from a rolling shutter.  The plane is
0.5 meter from the camera, the camera has a 40 degree field of view,
is operating at 30 frames per second.  From left to right, top to
bottom, the camera is shown rotating about the optical axis at
$\frac{1}{4}$, $\frac{1}{2}$, $\frac{3}{4}$ and $1$ revolution per
second, respectively.}
\end{figure}

However, most structure-from-motion algorithms (see e.g.
\cite{hartley99book,yimabook}) make the assumption that the exposure
of frames is instantaneous, and that every scanline is exposed at the
same time.  If either the camera is moving or the scene is not static
then the rolling shutter will induce geometric distortions when
compared with a camera equipped with a global shutter.  Consider, for
example, Figure \ref{checkboard} which shows the distortion present in
the image of a fronto-parallel checkerboard when a rolling-shutter
equipped camera is rotating about the optical axis.

The fact that every scanline is exposed at a different time will
lead to systematic biases in motion and structure estimates.Thus,
in considering the design of a robotic platform,which often by
necessity is in motion (for example a helicopter) one needs to
balance accuracy, cost and power consumption.  The goal of this
paper is to obviate this trade-off. By modeling the
rolling-shutter distortion we mitigate the reduction in accuracy
and present a framework for analyzing structure-from-motion
problems in rolling-shutter cameras

We define a general \emph{rolling-shutter constraint} on image points which is
shown to be true in the case of an arbitrarily moving rolling-shutter
camera.  Using this constraint we derive a projection equation for the
case of constant velocity fronto-parallel motion, which is exact when
the angular velocity about the optical axis is zero and furthermore
equivalent to a so-called ``crossed-slits'' camera.  The projection
equation is expressed in terms of the perspective projection plus a
correction term, and we show that bounds on its magnitude yield safe
regions in the space of depth vs. velocity where it is sufficient to
use pin-hole projection models.  For domains where the pin-hole model
is insufficient, we demonstrate a simple method to calibrate a rolling
shutter camera.  With new projection equations we re-derive the
optical flow equation, and we present experiments in simulation data.

Despite the prevalence of rolling-shutter cameras for consumer
electronics and web cameras, little has been done to address
issues related to structure-from-motion.  One of the few examples
which specifically model the rolling-shutter is the recent work by
Levoy et al. \cite{levoy04cvpr} who have constructed an array of
CMOS-equipped cameras for high speed videography.  The authors
propose to construct high framerate video by appropriate selection
of scanlines from an array of cameras, while compensating for the
rolling shutter effect. We are not aware of any research to
specifically address the rolling shutter phenomenon for
structure-from-motion.  However, such cameras are modeled in the
study of non-central cameras \cite{pless02omni}, that is, those
cameras which do not have a single effective viewpoint, or focus.
In fact, they are closely related to a specific type of camera
variously known as X-, crossed-, or two-slit cameras, discussed in
\cite{peleg03pami}, \cite{feldman03iccv} and \cite{pajdla02tr},
and generalized in \cite{yu94eccv}.  Instead of a single focus,
all imaged rays coincide with two slits.  We show here that under
certain circumstances, the rolling-shutter camera is a two-slit
camera.  However, the difference in focus between this paper and
most of the work on two-slit cameras is that two-slit cameras have
heretofore been simulated by selection of rows or columns from a
translating camera.  Therefore, calibration and motion estimation
can be done by traditional means using all the information in each
individual image, whereas we do not have that luxury.

In related works, Pless \cite{pless02omni} first described the
epipolar constraint and infinitessimal motion equations for arbitrary
non-central cameras, where cameras are represented by a subset of the
space of lines.  Pajdla \cite{pajdla02tr} has shown that the pushbroom
camera \cite{hartley94eccv}, which is approximates some satellite and
aerial imagers, is just one type of two-slit camera.  Latzel and
Tsotsos \cite{latzel01ip} have studied motion detection in the case of
a related sensor problem, namely interlaced cameras.

\section{Model of Rolling Shutter Cameras}

In this section we describe an equation for modeling cameras with
rolling-shutters.  When camera motion in the direction parallel to
the fronto-parallel plane is taking place we show how this effects
the projection and develop a projection equation independent of
time.  An interpretation of progressive scan as a so-called
``X-slit'' camera (meaning that instead of obeying a pin-hole
projection model the camera is modelled by two slits) results from
certain types of motion.

\begin{figure}[htb]
\centerline{
\includegraphics[scale=0.35]{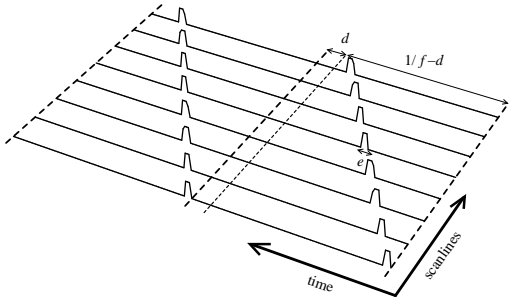}
}
\caption{\label{exposurefig}Rolling-shutter cameras expose scanline
sequentially; for some Firewire cameras the total exposure and the
scanning rate may be inversely proportional to the framerate.  The
scanline currently being scanned depends on the function
$v_{\text{cam}}(t)$, defined in \eref{rate}, which we assume is linear
in $t$. }
\end{figure}

We begin by describing the operation of the MDCS Firewire camera,
a specific rolling-shutter camera sold by Videre Design.  The
principles described here apply to other cameras with rolling
shutters, with only minor differences in some of the controllable
variables. The Firewire bus has a clock operating at 8Khz so the
camera can send no more than $8000$ packets per second.  The IIDC
specification for digital cameras \cite{IIDC} mandates that frames
sent over the bus should be equally spread out over these $8000$
packets.  The MDCS camera does not have an on-board buffer of
sufficient size to store a whole frame, and therefore it must send
data as soon as it is read from scanlines.  The scanlines are
sequentially exposed, read in, and immediately sent over the bus
such that the total time of exposure of one frame is inversely
proportional to the framerate.  The framerate($f$ frames/sec), the
exposure length of one scanline ($e$ $\mu$s), the rate at which
scanlines are exposed ($r$ rows/$\mu$s), and any delay between
frames ($d$$\mu$s)are the variables which control exposure of the
scanlines\footnote{The delay $d$ would normally be $0$ for IIDC
cameras with a rolling-shutter and without an on-board frame
buffer; for general cameras, though, this ought to be verified.}.
All of the $e$, $r$, and $d$ will in general be dependent in some
way on $f$, the framerate, and the camera.  An example of how a
set of scanlines is exposed is shown in Figure~\ref{exposurefig}.
We assume that the exposure time within a scanline is
instantaneous, i.e. that the peaks in Figure~\ref{exposurefig}
have zero width but integrate to some non-zero constant.  The
effect of $e$ being non-zero is motion blur within the scanline,
but has no geometric effects.

In a rolling-shutter camera each row of pixels is scanned in
sequential order over time to create an image.  We can formalize a
camera model by noting that each row corresponds to a different
instant in time, assuming that the exposure is instantaneous.  We
suppose that for a frame whose exposure starts at $t_0$, the row index
as a function of time is described by:
\begin{eqnarray}
\label{rate}
v_{\text{cam}}( \ssp t_0 \ssp + \ssp t \ssp )  & = &
r \, t - v_0 \, ,
\end{eqnarray}
where $r$ is the rate in rows per microsecond, and $v_0$ is the index
of the first exposed scanline.  The sign of $r$ depends on whether
scanning is top-to-bottom or bottom-to-top in the sensor.  For now let
us assume that $t_0 = 0$.

\begin{figure}[htb]
\centerline{
\begin{tabular}{cc}
{ \psfrag{XX}{\tiny$\hspace{1.5pt}X$}
\psfrag{QT}{\raisebox{1pt}{\tiny$\hspace{-6pt}q(t)$}}
\psfrag{VT}{\raisebox{-7pt}{\tiny$\hspace{-18pt}V(t)$}}
\includegraphics[scale=0.175]{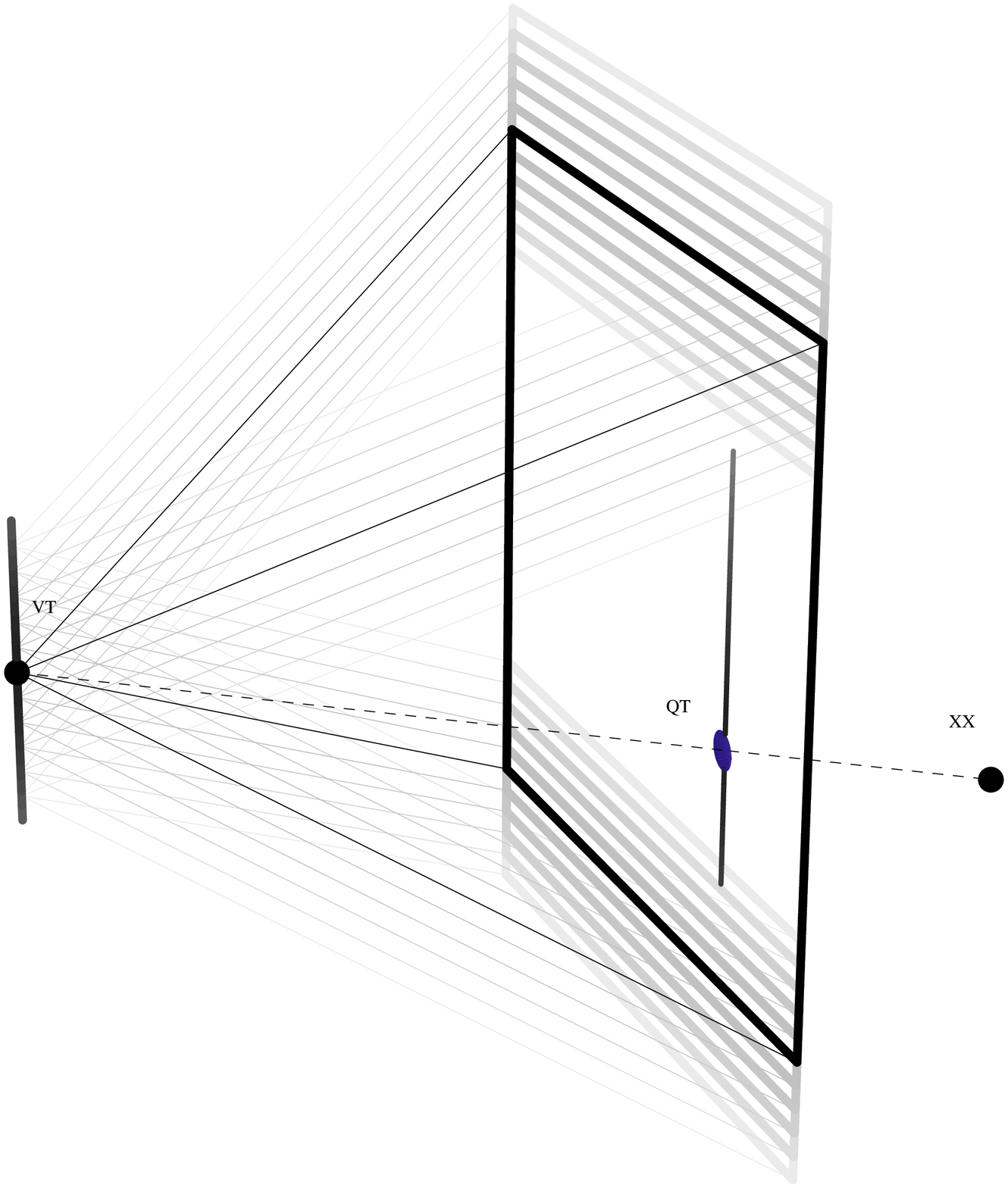} } \hspace{-20pt}&\hspace{-20pt}
{
\psfrag{VCAM}{\raisebox{-4pt}{\tiny$\hspace{-2pt}v_{\text{cam}}(t_c)$}}
\psfrag{QTC}{\raisebox{1pt}{\tiny$\hspace{-7pt}q(t_c)$}}
\includegraphics[scale=0.175]{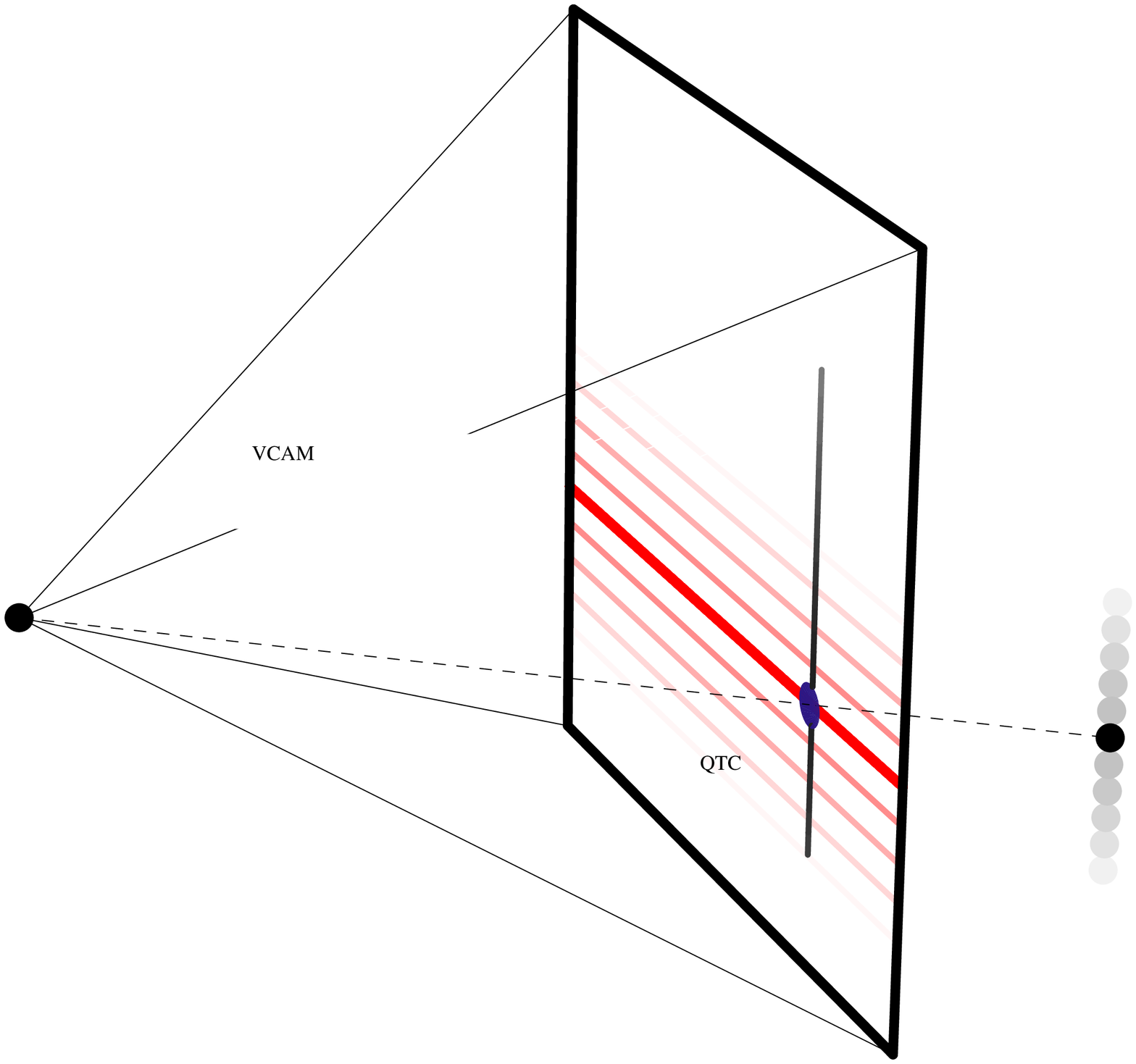} }
\end{tabular} }
\caption{\label{movcamfig}Left: the projection of a point in a moving
perspective camera generates some curve $q(t)$ in the image; Right: in
the reference frame of the camera, the point is apparently moving and
the is capture by the scanline at some $t_c$.}
\end{figure}

For an ideal perspective camera, the projection of a point as a
function of time is determined using the perspective camera equation:
\begin{eqnarray}
q(t) = \pi(\ssp P(t) \ssp X \ssp)
\end{eqnarray}
where $X = (x,y,z,1) \in \P^3$ represents, in homogeneous
coordinates, some static point in space; $P(t)$ is a camera matrix
as a function of time:
\begin{eqnarray*}
& P(t) \ssp = \ssp K \ssp \begin{bmatrix} R(t) & \!\!\!\bssp T(t)
\end{bmatrix} \,\text{ and }\, \pi(x,y,z,1) =
\left(\frac{x}{z},\frac{y}{z}\right)\!. & \label{Pdef}
\end{eqnarray*}
Here, $R(t)$ is an element of $\mathrm{SO}(3)$, the space of
orientation preserving rotations; $T(t) \in \R^3$ so that $V(t) =
-R(t) \ssp T(t)$ is the camera's viewpoint at time $t$ in a world
coordinate system; and finally $K$ is an upper-triangular calibration
matrix.  This configuration is shown in Figure \ref{movcamfig}.


Given that we have imaged some point $X$ at say $(u,v)$, it
\emph{must} be the case that:
\begin{eqnarray}
\begin{array}{rcccl}
\pi_y( \ssp P(t_c) \ssp X \ssp )
  & = & r \, t_c - v_0 & = & v \
\end{array} \label{progscaneqn}
\end{eqnarray}
for some time $t_c$, and where $\pi_y$ represents the $y$-component of
the projection $\pi$.  Equation \eref{progscaneqn} is the general
\emph{rolling-shutter constraint}.  The left hand side describes the
curve of the projected point over time, while the right hand side of
the equation represent the scanline as a function of time.  The
scanline captures the curve of the projected point at some time
instance $t_c$.

If the camera is stationary, i.e. $P(t)$ is independent of $t$,
then the left hand side of equation \eref{progscaneqn} is
independent of time so the image of $X$ is independent of $t_c$
and the resulting projection is the usual perspective one.
However, in general the camera will be moving and the resulting
projection will depend on $t_c$ in some way.  Under certain
assumptions on $P(t)$ we can solve analytically for $t_c$,
substitute it into the left hand side of \eref{progscaneqn} and
obtain a projection formula.  The goal of the next section is to
derive specific projection formulas in common situations.

\subsection{Fronto-parallel motion}

Suppose that a calibrated camera undergoes a constant angular velocity $\omega$
and a linear velocity $v$.  A linear approximation of $P(t_c)$ about
$t_c = 0$ is given by:
\begin{eqnarray*}
P( t_c ) & \approx & K
\begin{bmatrix}
\, ( I + t_c \, \hat{\omega} ) R(0) & T(0) + v \, t_c \,
\end{bmatrix} . \label{Ptapprox}
\end{eqnarray*}
When $\omega = 0$, i.e., when there is no angular velocity, the
linearization is exact; otherwise it is an approximation.  Let us
analyze the case of constant velocity fronto-parallel motion; that is,
\begin{eqnarray*}
& v \, = \, [v_x,v_y,0]^T \, \text{ and } \, \omega =
[0,0,\omega_z]^T . &
\end{eqnarray*}
Substituting the approximation of \eref{Ptapprox} into equation
\eref{progscaneqn}, the rolling-shutter constraint, yields an equation
linear in $t_c$.  Assuming without loss of generality that $R(0) = I$,
$T(0) = [0,0,0]^T$, and $K = I$, then after substituting the solution
to $t_c$ back into equation $q(t_c)$, we obtain:
\begin{eqnarray}
\!\!\!\!\!q_{\text{rolling}\atop\text{shutter}}\! = \, q(t_c)  \!\!\!\! & = & \!\!\!\!
\begin{bmatrix}
\frac{x}{z} + \frac{y + v_0 z}{( r \ssp z - v_y - w_z \ssp x )} \frac{( v_x - w_z y )}{z} \\
\frac{y}{z} + \frac{y + v_0 z}{( r \ssp z - v_y - w_z \ssp x )} \frac{( v_y + w_z x )}{z} \\
\end{bmatrix}\, \label{rollshutterproj} \\
& = & \!\!\!\!  q(0) \;\ssp + \;\,\ssp q_{\text{\tiny{}correction}} \ssp .
\nonumber
\end{eqnarray}
\emph{This is the image of the point $(x,y,z)$ in the frame of a
rolling-shutter camera starting at $t=0$}.  If $\omega = 0$, this
equation is exact.  Therefore, we arrive at a projection equation
for a moving camera with a rolling-shutter which is independent of
time. We write the projection equation so that it is more obvious
what the effects of the rolling-shutter are, namely that the
resulting projection equals the ideal perspective projection plus
a correction term which is proportional to the optical flow
$[\frac{v_x-w_z y}{z},\frac{v_y+w_z x}{z}]^T$ of a perspective
camera.  Note that as the rate of scan, $r$, goes to $\infty$, the
limit of the correction term becomes $[0,0]^T$, corresponding to a
camera with a global instantaneous shutter.

\begin{figure*}[htb]
\centerline{
\begin{tabular}{ccc}
\includegraphics[scale=0.107]{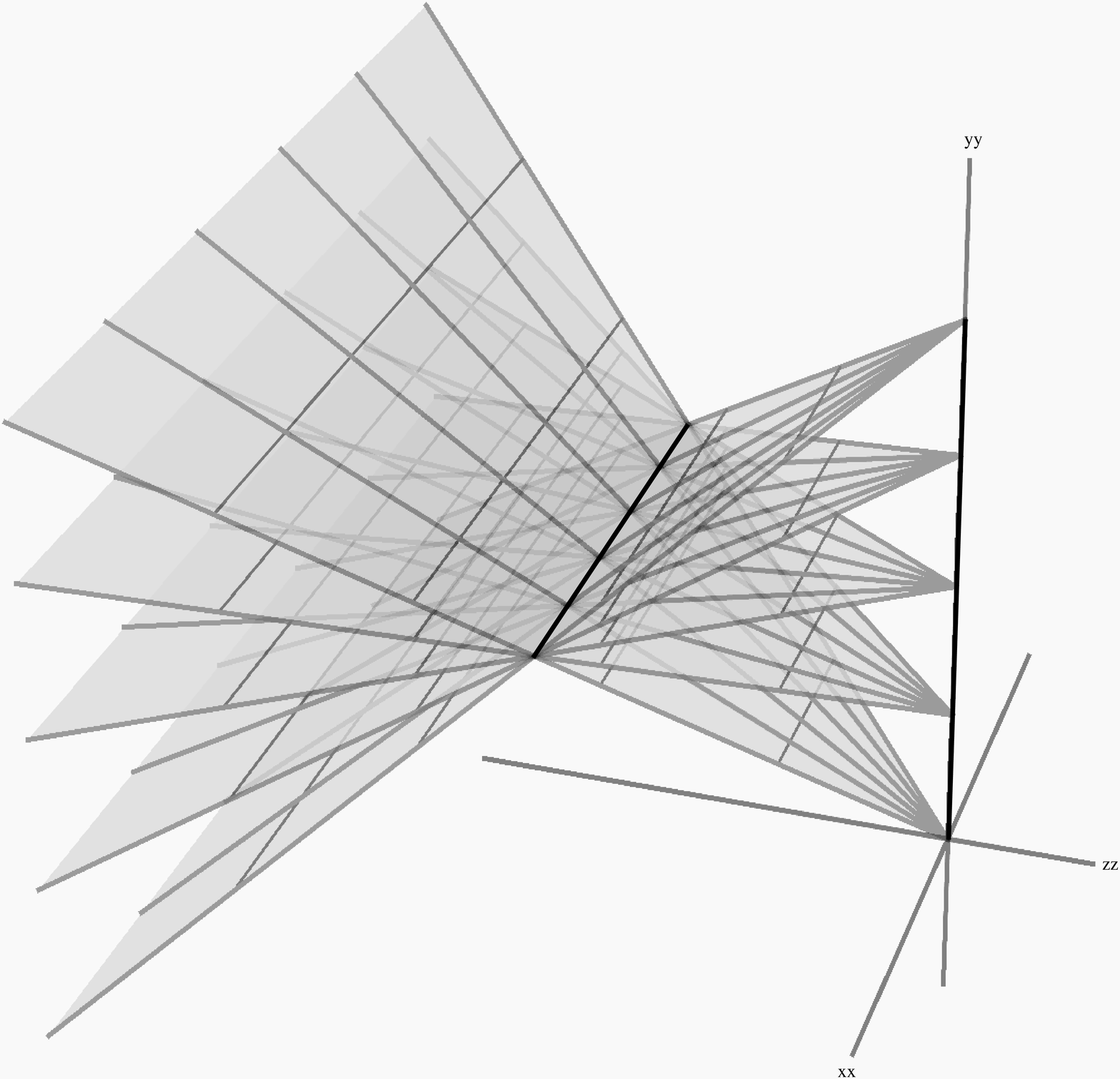} &
\includegraphics[scale=0.107]{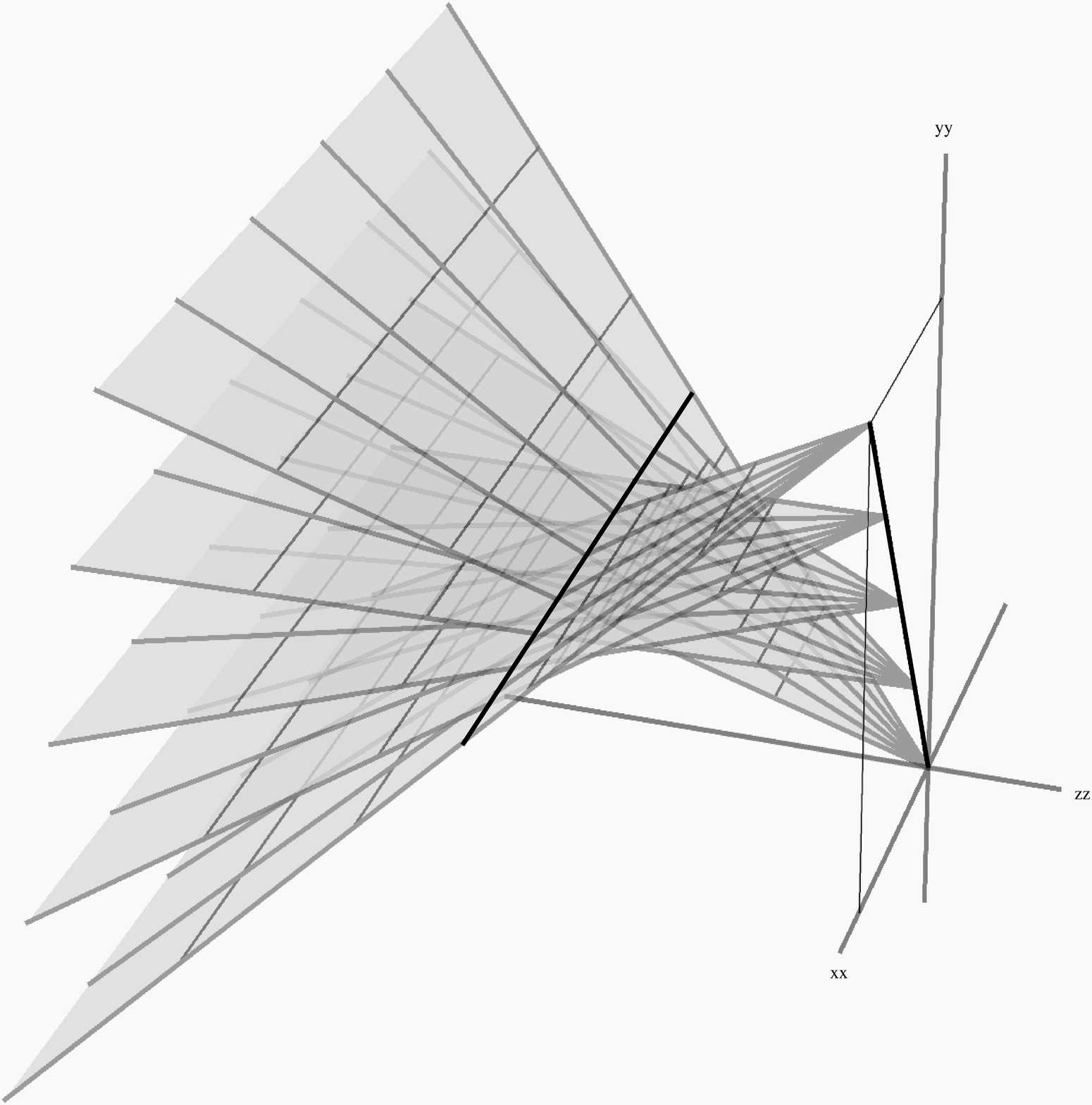} &
\includegraphics[scale=0.102]{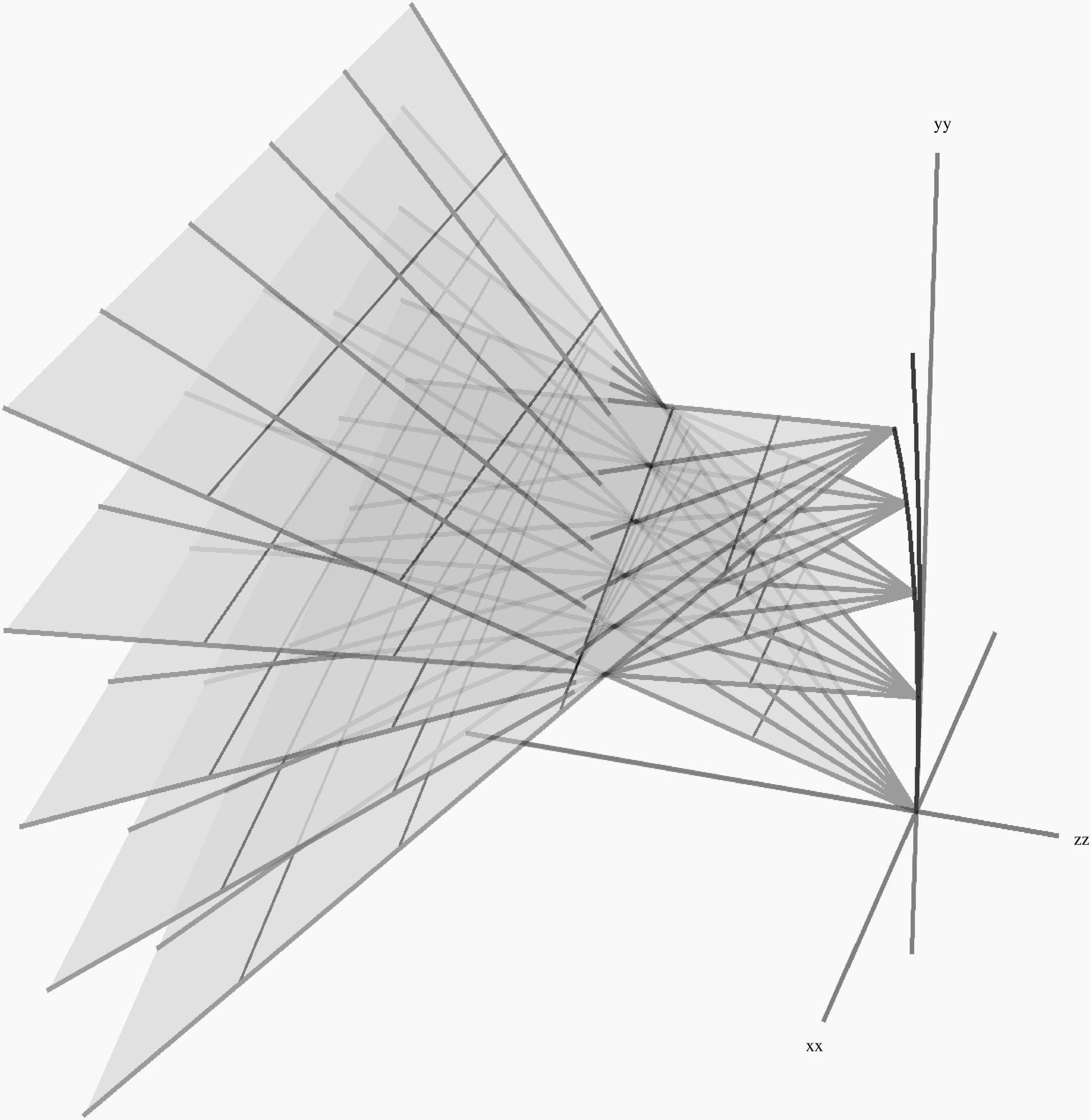}
\end{tabular}}
\caption{\label{figXslit2} Left: When only $v_y \neq 0$, the camera
becomes an orthogonal crossed-slit camera; middle:
when $v_x, v_y \neq 0$ are non-zero, the camera is
a general crossed-slit camera; and right: when $\omega \neq 0$,
the approximating camera is no longer a crossed-slit camera.}
\end{figure*}

Thus, what a camera with a rolling-shutter becomes is a camera whose
projection geometry is parameterized by its velocity.  When
stationary, the camera obeys the pinhole projection model.  However,
when undergoing linear motion with constant velocity, one can show
that the camera becomes a crossed-slits camera.  For any point $(u,v)$
we may invert \eref{rollshutterproj} up to an unknown $z$ and
demonstrate that all such inverse image coincide with two lines: the
first being the line through $(0,0,0)$ and $(v_x,v_y,0)$; and the
second through the points $(\pm 1,\frac{v_0 v_y}{r},\frac{v_y}{r})$.
As $v_x$ and $v_y$ approach $0$, these two slits coincide at
$(0,0,0)$.  The case when $\omega_z = v_x = 0$ is shown in the left
part of Figure \ref{figXslit2}; in the middle we show the case when
$v_x$ and $v_y$ are arbitrary, and $\omega_z = 0$; finally in the
right part of Figure \ref{figXslit2} we show that the resulting
geometry when $\omega_z \neq 0$, in which case the camera no longer
obeys the crossed-slits camera model.

\subsection{Domain of applicability}
\label{domainapp} 
\begin{figure}[htb]
\centerline{\includegraphics[scale=0.3]{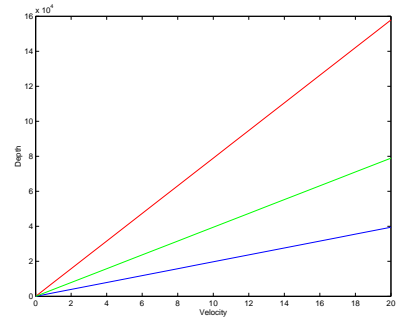}}
\caption{\label{correctionTerm} The limit line for a 640 x 480 image
for different framerates. Blue represents 3.75fps, green is 7.5fps,
and red is 15fps.}
\end{figure}

Under what condition is the perspective model a
sufficient model for rolling-shutter cameras?  If error in feature
location is on the order of one pixel, then we would expect
distortions due to the rolling-shutter to be subsumed in noise.
Thus, the maximum value that the correction term can obtain should
be less than one pixel length in order for a perspective camera
model to be an adequate representation of the rolling shutter
camera.  The correction term varies as a function of velocity and
depth $z$.  In order to determine when this correction term needs
to be factored in, and a perspective model is no longer suitable,
these parameters must be evaluated in light of the fixed
parameters of the camera and the shutter rate $r$.  Given that
$s_{\alpha}$ is the length of a pixel, it can be seen that:
\begin{eqnarray}
\frac{1}{s_{\alpha}r} v_y < z
\end{eqnarray}
must be satisfied in order for the rolling shutter effects to be
inconsequential.  However, when this inequality does not hold, then
the rolling shutter projection model should be used or there will be a
bias in estimations.  This creates a limit line, as a function of
$v_y$ and $z$, based on the shutter rate and $s_{\alpha}$, as shown in
Figure\ref{correctionTerm}.  When the values are such that the camera
is operating well above this line, a rolling shutter model should be
used.

\begin{table*}[htb]
\centerline{
\begin{tabular}{|l|l|l|} \hline
Case & Solution to $t_c$ & Projection equation \\
\hline
\begin{minipage}{2.2in}
\begin{flushleft}
\vspace{2pt}
Constant velocities $v =[ v_x, v_y, 0]$, $w = [ 0, 0, w_z]$, and
linear approximation of $P(t)$.
\vspace{2pt}
\end{flushleft}
\end{minipage} &
\begin{minipage}{1.5in}
\begin{flushleft}
Results in a linear equation which can be solved for $t_c$.
\end{flushleft}
\end{minipage} &
\begin{minipage}{2.6in}
\begin{flushleft}
Equation \eref{rollshutterproj}.
\end{flushleft}
\end{minipage}
\\
\hline
\begin{minipage}{2.2in}
\begin{flushleft}
Constant velocities $v = [ 0, 0, v_z]$, $w = [ 0, 0, 0]$, and linear
approximation of $P(t)$.
\end{flushleft}
\end{minipage} &
\begin{minipage}{1.5in}
\begin{flushleft}
Results in a quadratic equation for $t_c$, choose smallest positive
root.
\end{flushleft}
\end{minipage} &
\begin{minipage}{2.7in}
\begin{flushleft}
\vspace{2pt}
The smallest positive root of $t_c$ when sub\-stituted into $q(t_c)$
yields an analytic equation for the projection without $t_c$.
\vspace{2pt}
\end{flushleft}
\end{minipage}
\\
\hline
\begin{minipage}{2.2in}
\begin{flushleft}
\vspace{2pt}
Constant velocities $v = [ v_x, v_y, v_z]$, $w = [ w_x, w_y, w_z]$,
and linear approximation of $P(t)$.
\vspace{2pt}
\end{flushleft}
\end{minipage} &
\begin{minipage}{1.5in}
\begin{flushleft}
Also results in a quadratic equation in $t_c$.
\end{flushleft}
\end{minipage} &
\begin{minipage}{2.6in}
\begin{flushleft}
Using the solution for $t_c$ in the general projection $q(t_c)$ leads
to a projection model is analytic and independent of $t_c$.
\end{flushleft}
\end{minipage}
\\
\hline
\begin{minipage}{2.2in}
\begin{flushleft}
\vspace{2pt}
Constant velocities $v = [v_x, v_y, v_z]$, $w = [w_x, w_y, w_z]$,
without linear approximation of $P(t)$.
\vspace{2pt}
\end{flushleft}
\end{minipage} &
\begin{minipage}{1.5in}
\begin{flushleft}
Nonlinear equation for $t_c$; can be solved with a root
finder (when $(x,y,z)$ are \emph{known}).
\end{flushleft}
\end{minipage} &
\begin{minipage}{2.6in}
\begin{flushleft}
The projection equation $q(t_c)$ is nonlinear in $t_c$.
\end{flushleft}
\end{minipage} \\
\hline
\end{tabular}}
\caption{\label{motionTypes}
Different assumptions on the motion, and our
choice in whether to linearize $P(t)$ lead to
different types of solutions for $t_c$ and
the projection $q(t_c)$.}
\end{table*}

\subsection{General rotation and translation}
Generalizing the motion for any $T = [ v_x v_y v_z]^T$ and an
arbitrary rotation matrix $R$ we can see the relation of the
rolling-shutter camera projection to $t_c$ for different. We
discussed the linearity of  the fronto-parallel case previously
and can be see in \eref{Ptapprox}. When there is a non-zero
velocity along the principle axis of the camera, $v_z$, the
relation now becomes quadratic. Using a root solver, we can find a
solution of $t_c$ to make a rational projection equation.
Different types of motion and the relation of the projection
function to $t_c$ follow from the rolling-shutter rate \eref{rate}
and \eref{Pdef}.  This relation and its effects on the projection
are shown in Figure \ref{motionTypes}.

The linearized $P(t)$ and the resulting projection equation for
rolling shutter camera can be used given any type of camera motion.
In certain cases $t_c$ is a solution to a linear equation and in other
cases, $t_c$ is a solution to a quadratic.  In both instances $t_c$
can be found either directly or using a root solver and then used in
the general projection equation to find a projection model of the
rolling shutter that is independent of $t_c$.  The resulting
projection model will only be dependent on the velocity of the camera,
the internal parameters, and the shutter rate.

\section{Calibration}

In this section we discuss the calibration of a Firewire camera with
rolling shutter, namely the MDCS camera from Videre design.  Our goal
is to determine the constant $r$, the rate of scan, and determine if
$d$, the delay between frames, is zero.  If we find that $r = n_r/f$,
where $n_r$ is the number of rows and $f$ is the framerate, then we
can conclude that $d = 0$.

\begin{figure}[htb]
\label{led}
\centerline{\includegraphics[scale=0.5]{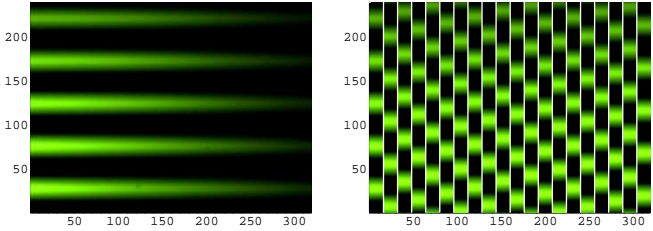}}
\caption{ Left: an example image taken during calibration; right:
an example of the spatio-temporal image $I(y,t)$ used to estimate
the rate $r$.}
\end{figure}

Since the camera exposes only several neighboring scanlines at a
time, we can indirectly measure $r$ by placing an LED flashing at
known frequency in front of the camera and measuring the peak
frequency of vertical slices of the image in the frequency domain.
In a simple experiment we placed the camera in front of a green
LED, removing its lens to ensure that all pixels are exposed by
the LED, and draped a dark cloth over the setup so as to eliminate
ambient light.  An example of a captured image is shown in Figure
\ref{led}; we were unable to achieve even distribution of exposure
over the length of the scanline.  For any choice of framerate and
frequency, we selected and summed a subset of the columns from
each image, yielding for any one time $t_i$ the vector $I(y,t_i)$,
thereby constructing an image $I(y,t)$ for each experiment whose
independent variable is the pair of framerate and frequency.  The
duration of exposure, which is a configurable parameter, affects
the width of the horizontal bars in Figure \ref{led}, though it
does not affect their period.  To estimate the frequency we take
the two-dimensional Fourier transform of $I(y,t)$, obtaining
$\hat{I}(\nu,\omega)$ which we marginalize over $\omega$ to yield
$\tilde{I}(\nu)$.  The estimate of $r$ is the location of the
highest frequency peak (after high frequency components are
ignored because the image is not a sine wave).  Thus, using the
image based frequency of the light flashes in conjunction with the
actual frequency of the LED flashes from the pulse generator, we
were able to approximate the rate of the rolling shutter for the
camera.  It can be seen in Figure~\ref{CalTable} that it is
sufficient to model our camera as having no delay
\footnote{Frequency of they LED frequency (pulse generator
frequency) was varied from 2.5Hz to 103Hz. Not all framerates
tested the complete span of LED frequencies}.

\begin{table}[htb]
\label{CalTable}
\centerline{
\begin{tabular}{|l|c|r|}\hline
FPS & calibrated time & ideal time\\
& (sec/row) & (sec/row)\\\hline
3.75 & 0.00110 $\pm$ 0.00050 & 0.00110\\\hline 7.5 & 0.00063 $\pm$ 0.00050 & 0.00056\\
\hline 15 & 0.00029 $\pm$ 0.00050 & 0.00028\\ \hline
\end{tabular}}
\caption{The calibrated time to scan a row in relation to the
ideal time to scan a row given the framerate. Note that the
accuracy of the pulse generator frequency is taking into account
in the offset number}
\end{table}

\begin{figure}[htb]
\centerline{\includegraphics[scale=0.4]{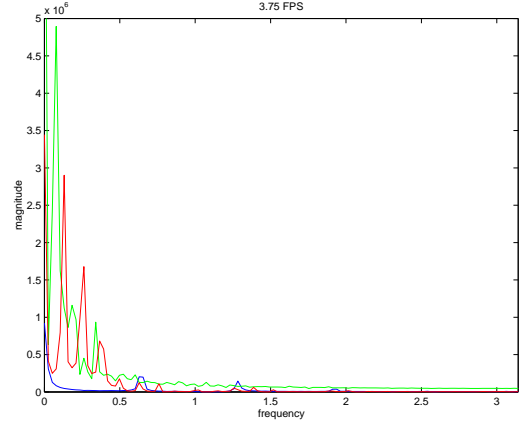}}
\caption{\label{frequency}The positive frequencies of the
marginalized 2D Fourier Transform of the LED capture image for
framerate of 3.75 FPS. The blue shows the results using a pulse
generator frequency(LED frequency) of 103Hz. The red shows the
results for a pulse generator frequency of 20Hz. The green shows
the results for pulse generator frequency of 11Hz. }
\end{figure}

\subsection{Optical flow equations}

To determine the optical flow equations for
rolling-shutter cameras, we make the assumption
that the parameterized camera can be sampled
infinitely often in time.  Furthermore,
for ease of presentation we will assume from now on
that $v_0 = 0$; in general this may induce some
$t_c < 0$.  To determine the flow we solve the
rolling-shutter constraint using the camera
$P(t_0 + t_c)$, and having found and substituted
$t_c$ as well as expressions for $x$ and $y$ in terms of the
image coordinates and $z$, we differentiate with respect to $t_0$ and
evaluate at $t_0 = 0$.  For the case of fronto-parallel
motion, i.e. $v = (v_x,v_y,0)^T$ and $\omega =
(0,0,\omega_z)$, and linearization of $t_0$, this
procedure yields the following optical flow:
\begin{eqnarray}
\dot{q}(u,v) & = &
\frac{r \ssp z}{v \ssp v_x w_z + r \ssp z (r - \dot{v}_p)}
\begin{pmatrix}
r \ssp \dot{u}_p + \omega_z v \ssp \dot{v}_p  \\
r \ssp \dot{v}_p + \omega_z v \ssp \dot{u}_p
\end{pmatrix},
\label{rsopticalflow}
\end{eqnarray}
where $(\dot{u}_p,\dot{v}_p)$ is the optical flow
for perspective cameras in the case of fronto-parallel
motion:
\begin{eqnarray*}
(\dot{u}_p,\dot{v}_p) & = &
\begin{bmatrix}
\frac{v_x}{z} - \omega_z v, &
\frac{v_y}{z} + \omega_z u
\end{bmatrix}.
\end{eqnarray*}
In the case of general motion and linearization of $P(t)$, the
resulting equation for optical flow remains analytic but is
significantly more complicated.

\section{Simulation}

\begin{figure*}[htb]
\centerline{
\psfrag{DD}{\hspace{-35pt}\footnotesize{}Velocity: $1.875$ km/h.}
\psfrag{CC}{\hspace{-35pt}\footnotesize{}Velocity: $3.75$ km/h.}
\psfrag{BB}{\hspace{-35pt}\footnotesize{}Velocity: $5.625$ km/h.}
\psfrag{AA}{\hspace{-35pt}\footnotesize{}Velocity: $7.5$ km/h.}
\psfrag{PIXERR}{\hspace{-25pt}\tiny{}Simulated error (pixels)}
\psfrag{QA}{\rotatebox{90}
{\hspace{-33pt}\begin{minipage}{1in}\vspace{-3pt}\begin{center}\tiny{}Reprojection error\\ (for comparison only)\end{center}\end{minipage}}}
\psfrag{QB}{\rotatebox{90}
{\hspace{-20pt}\begin{minipage}{1in}\vspace{-3pt}\tiny{}Mean rotation
\\error (degrees)\end{minipage}}}
\psfrag{QC}{\rotatebox{90}
{\hspace{-33pt}\begin{minipage}{1in}\vspace{-3pt}\tiny{}Mean error in
direction\\ of translation (degrees)\end{minipage}}}
\includegraphics[scale=0.6]{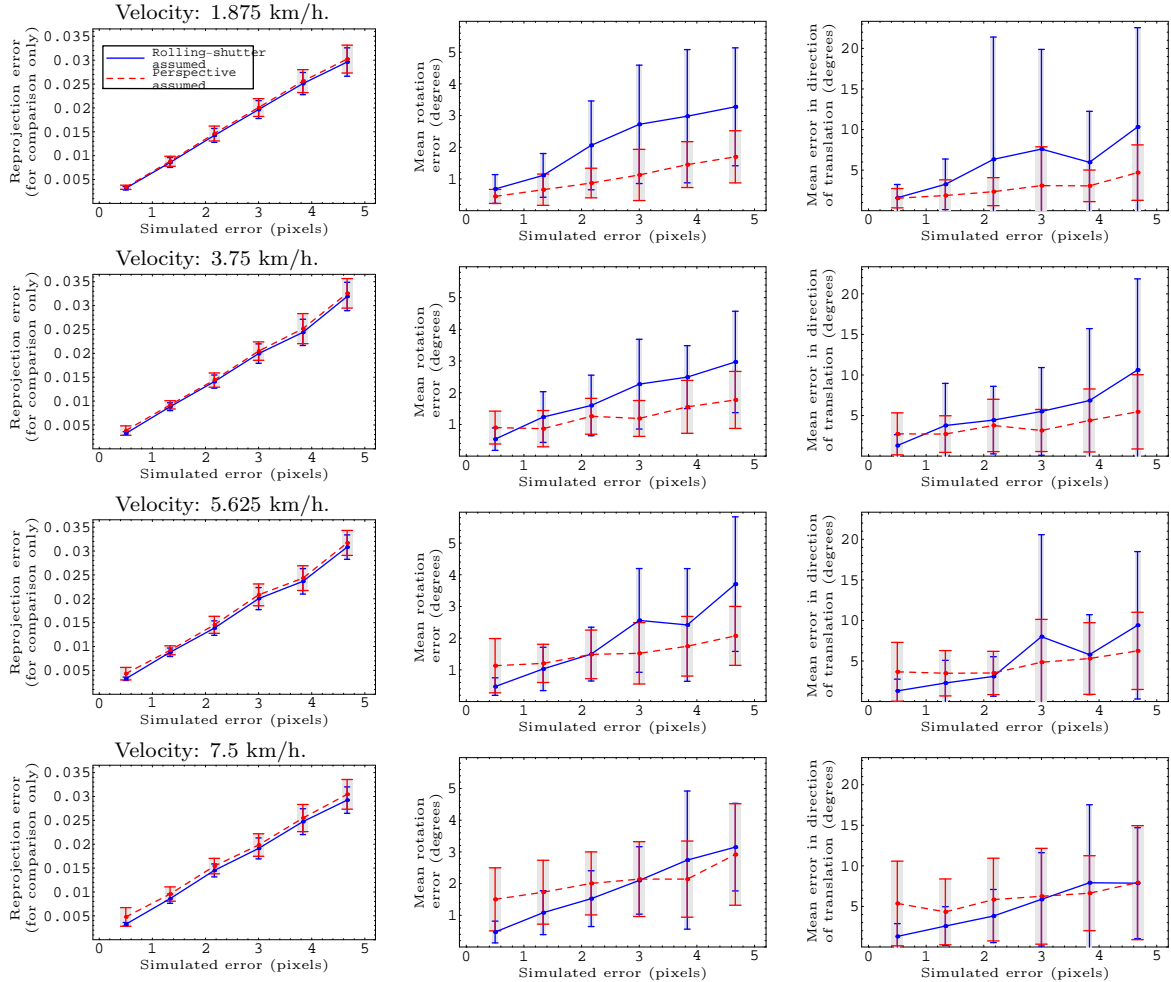}}
\caption{\label{simfigs} Plot of reprojection errors, errors in rotation and translation,
and for varying velocities.}
\end{figure*}

In section \ref{domainapp} we determined those conditions under which
it may be wise to take into account distortions caused by a rolling
shutter.  In this section we test this claim in simulations under
varying speeds and noise conditions.  We randomly generate a set of
$100$ points and project these points into two cameras according to
the rolling-shutter camera model for certain $v_y \neq 0$, all others
zero.  The cameras are placed randomly, but on the order of $10$
meters from the point cloud.  Then, assuming a camera sampling at $15$
frames per second, with scanning rate inversely proportional to the
framerate, we let $v_y = 1.875$, $3.75$, $5.625$ and $7.5$ km/h.  For
each velocity we perform experiments with pixel noises $\sigma = 0.5$,
$1.33$, $2.16$, $3$, $3.83$ and $4.66$ pixels.

For each instantiation of the experiment, we perform bundle adjustment
on the image points using a projection model incorporating the rolling
shutter (blue, solid lines) and the perspective projection model (red,
dashed lines).  The errors, plotted in Figure \ref{simfigs}, are:
left: reprojection error, middle: the error of the estimate $\hat{R}$
of rotation when compared with the true $R_0$, measured as $\| \log
R_0^{-1} \hat{R} \|$; and, right: the error of the estimate $\hat{t}$
of translation when compared with the true $t_0$, measured as
$\cos^{-1} \hat{t}^T t_0 / ( \| \hat{t} \| \, \| t_0 \| )$.

For high velocities or low noise we find that by compensating for the
rolling shutter, one is able to do better than if one were to ignore
the effect.  However, for lower velocities and higher noise, the noise
effectively becomes louder, comparatively, than the deviation due to
the rolling-shutter.

\section{Conclusion}

 Despite the prevalence of rolling-shutter cameras little has been
done to address the issues related to structure-from-motion.  Since
these cameras are becoming more common, the effects due to the shutter
rate must be taken into account in vision systems.  We have presented
an analysis of the projective geometry of the rolling-shutter
camera. In turn, we have derived the projection equation for such a
camera and demonstrated that under certain conditions on the motion
and internal parameters, using this flow equation can increase
accuracy when doing structure-from-motion.

The effect of the rolling-shutter becomes important when either the
camera or objects in the scene are moving, as the shutter rate can
produce distortion if not accounted for.  The modeling we have
presented may play an important role for incorporating rolling-shutter
cameras into moving systems, such as UAV's and ground robots. Given
the improved accuracy in structure-from-motion, this can aid in scene
recovery and navigation of systems operating in dynamic worlds.
Future work includes testing these models on UAV's for recovery of
scene structure.


\begin{thebibliography}{10}\setlength{\itemsep}{-1ex}\small

\bibitem{IIDC}
{1394 Trade Association}.
\newblock {\em {IIDC} 1394-based {D}igital {C}amera {S}pecification
  {V}er.1.30}.
\newblock July 2000.

\bibitem{feldman03iccv}
D.~Feldman, T.~Pajdla, and D.~Weinshall.
\newblock On the epipolar geometry of the crossed-slits projection.
\newblock In {\em Proceedings of International Conference on Computer Vision},
  pages 988 -- 995, October 2003.

\bibitem{hartley94eccv}
R.~Hartley and R.~Gupta.
\newblock Linear pushbroom cameras.
\newblock In {\em Proceedings of European Conference on Computer Vision}, 1994.

\bibitem{hartley99book}
R.~Hartley and A.~Zisserman.
\newblock {\em Multiple View Geometry}.
\newblock Cambridge Univ. Press, 2000.

\bibitem{latzel01ip}
M.~Latzel and J.~K. Tsotso.
\newblock A robust motion detection and estimation filter for video signals.
\newblock In {\em Proceedings of Image Processing}, pages 381 -- 394, October
  2001.

\bibitem{yimabook}
Y.~Ma, S.~Soatto, J.~Ko{\v{s}}eck{\'{a}}, and S.~Sastry.
\newblock {\em An Invitiation to 3D Vision: From Images to Geometric Models}.
\newblock Springer Verlag, New York, 2003.

\bibitem{pajdla02tr}
T.~Pajdla.
\newblock Geometry of two-slit camera.
\newblock Czech Technical University, Technical report CTU-CMP-2002-02, 2002.

\bibitem{pless02omni}
R.~Pless.
\newblock Discrete and differential two-view constraints for general imaging
  systems.
\newblock In {\em Proceedings of the 3rd Workshop on Omnidirectional Vision},
  June 2002.

\bibitem{wany03ted}
M.~W{\"a}ny and G.~P. Israel.
\newblock {CMOS} image sensor with {NMOS}-only global shutter and enhanced
  responsivity.
\newblock {\em Trans. on Electron Devices}, 50(1), January 2003.

\bibitem{levoy04cvpr}
B.~Wilburn, N.~Joshi, V.~Vaish, M.~Levoy, and M.~Horowitz.
\newblock High speed video using a dense camera array.
\newblock In {\em Proceedings of Computer Vision and Pattern Recognition},
  2004.

\bibitem{yu94eccv}
J.~Yu and L.~McMillan.
\newblock General linear cameras.
\newblock In {\em Proceedings of European Conference on Computer Vision}, 2004.

\bibitem{peleg03pami}
A.~Zomet, D.~Feldman, S.~Peleg, and D.~Weinshall.
\newblock Mosaicing new views: the crossed-slits projection.
\newblock {\em Trans. on Pattern Analysis and Machine Intelligence}, 25(6),
  June 2003.

\end{thebibliography}

\end{document}